\documentclass[letterpaper, 10 pt, conference]{ieeeconf}  % Comment this line out if you need a4paper

\IEEEoverridecommandlockouts                              % This command is only needed if 
\usepackage{graphics} % for pdf, bitmapped graphics files
\usepackage{epsfig} % for postscript graphics files
\usepackage{mathptmx} % assumes new font selection scheme installed
\usepackage{times} % assumes new font selection scheme installed
\usepackage{amsmath} % assumes amsmath package installed
\usepackage{amssymb}  % assumes amsmath package installed
\usepackage{multirow}
\usepackage{color,soul}
\usepackage{algorithm}
\usepackage{siunitx}
\usepackage{algpseudocode}

\usepackage{hyperref}
\usepackage[comma,numbers]{natbib}

\title{\LARGE \bf
High-Speed and Impact Resilient Teleoperation of Humanoid Robots
}

\author{Sylvain Bertrand$^{1}$,
Luigi Penco$^{1}$, 
Dexton Anderson$^{1}$,
Duncan Calvert$^{1,2}$,
Valentine Roy$^{1}$,\\
Stephen McCrory$^{1,2}$,
Khizar Mohammed$^{3}$,
Sebastian Sanchez$^{3}$,
Will Griffith$^{3}$,
Steve Morfey$^{1,3}$,\\
Alexis Maslyczyk$^{1,3}$,
Achintya Mohan$^{3}$,
Cody Castello$^{1}$,
Bingyin Ma$^{3}$,
Kartik Suryavanshi$^{3}$,
Patrick Dills$^{3}$,\\
Jerry Pratt$^{3}$,
Victor Ragusila$^{3}$,
Brandon Shrewsbury$^{3}$,
and Robert Griffin$^{1,2}$% <-this % stops a space
\thanks{This work was funded through ONR Grant N00014-19-1-2023.}% <-this % stops a space
\thanks{$^{1}$At the time of this work, the authors were with the Florida Institute for Human and Machine Cognition, 40 S Alcaniz St, Pensacola, FL 32502, United States}%
\thanks{$^{2}$At the time of this work, the authors were with the University of West Florida, 11000 University Pkwy, Pensacola, FL 32514, United States}%
\thanks{$^{3}$At the time of this work, the authors were with Boardwalk Robotics, 417 E Zaragoza St, Pensacola, FL 32502, United States}%
\thanks{Email : \url{sbertrand@ihmc.org}, \url{lpenco@ihmc.org}, \url{brandon.shrewsbury@boardwalkrobotics.com}, \url{rgriffin@ihmc.org}
}
\thanks{Thanks to David Oh for his contributions on this work.}} 

\begin{document}

\maketitle
\thispagestyle{empty}
\pagestyle{empty}

\begin{abstract}
Teleoperation of humanoid robots has long been a challenging domain, necessitating advances in both hardware and software to achieve seamless and intuitive control. This paper presents an integrated solution based on several elements: calibration-free motion capture and retargeting, low-latency fast whole-body kinematics streaming toolbox and high-bandwidth cycloidal actuators. Our motion retargeting approach stands out for its simplicity, requiring only 7 IMUs to generate full-body references for the robot. The kinematics streaming toolbox, ensures real-time, responsive control of the robot's movements, significantly reducing latency and enhancing operational efficiency. Additionally, the use of cycloidal actuators makes it possible to withstand high speeds and impacts with the environment. Together, these approaches contribute to a teleoperation framework that offers unprecedented performance. Experimental results on the humanoid robot Nadia demonstrate the effectiveness of the integrated system (\url{https://youtu.be/F6dqCauGPEM}).
\end{abstract}

\section{Introduction}
\label{introduction}
Despite its potential and the flexibility offered by intuitive human-like movement generation and the integration of human decision-making, the teleoperation of humanoid robots still encounters significant challenges, exposing considerable limitations in their operational efficiency.

One significant challenge in teleoperation is achieving high transparency. This refers to the ability to control the robot with the same bandwidth and speed as human motion, without any noticeable delay. Attaining high transparency necessitates a combination of sophisticated solutions that ensure the real-time streaming of user commands to the robot while maintaining the robot's balance.

Some research has focused on using motion anticipation to minimize delays and achieve synchronization. For example, \cite{dallard2023} aims to imitate and predict an operator's intentions at two levels: first, by anticipating upper body motions through hand movement predictions using polynomial interpolation that aligns with a minimum jerk model; second, by evaluating and predicting walking pace and step locations using a recurrent neural network. 
Another technique uses probabilistic motion primitives to forecast user movements \cite{penco2023}. This method compensates for round-trip delays in visual feedback by predicting the user's future states and adjusting the robot's actions accordingly. These techniques are promising for reducing the lag in teleoperation systems. However, these anticipation methods have not yet been validated into scenarios demanding high-speed motion.
\begin{figure}
    \centering
        \includegraphics[width=1.0\columnwidth]{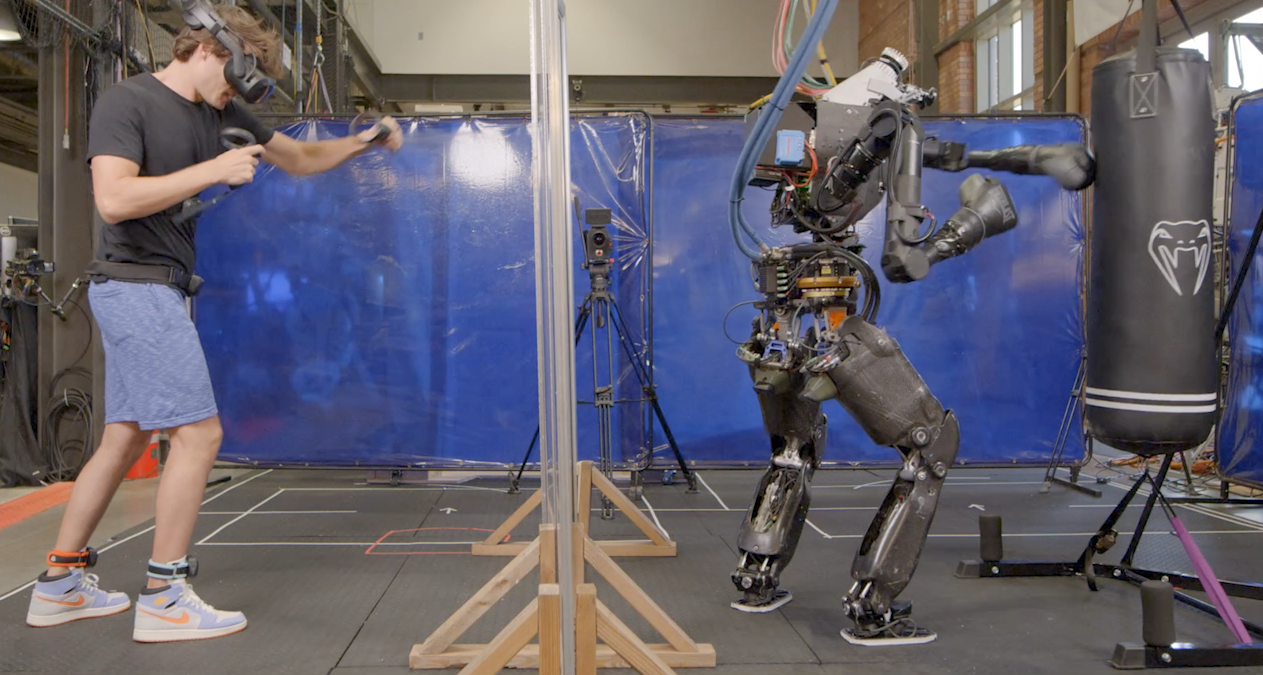}
    \caption{The humanoid robot Nadia being teleoperated by a human operator equipped with a VR headset, four VR trackers, and two VR controllers. Thanks to our kinematics streaming framework and the integration of cycloidal actuators, Nadia can punch a 100 lbs punching bag at human speed.}
    \label{fig:intro}
\end{figure}

Efforts by other researchers have focused on synchronizing human and robot dynamics through bilateral teleoperation techniques, which directly couple human and robot motion through haptic feedback \cite{teleopSurvey}. For example, Ramos \textit{et al.} \cite{ramos2018, colin2023} explore the direct coupling of human and humanoid dynamics using a human-machine interface, achieving synchronized locomotion. Ishiguro \textit{et al.} \cite{ishiguro2020} explored a different approach by employing a full-body exoskeleton cockpit for bilateral teleoperation, aiming to align the operator's and the robot's movements seamlessly. These research approaches hold significant promise for enhancing teleoperation, particularly in dynamically aligning human and robot motions. However, they come with certain trade-offs. For instance, they are highly sensitive to network delays, which can impact their effectiveness in real-time applications and require the use of large, complex structures on the user side, which can be cumbersome and require maintenance.

Some recent efforts have aimed to simplify the user-side setup. For instance, methods like shadowing \cite{fu2024humanplus, he2024learning} allow humanoid robots to mimic human body and hand movements in real-time using only RGB cameras, based on pretrained policies.
However, these methods rely on extensive training with large human datasets and have yet to demonstrate their effectiveness in high-speed, synchronized teleoperation scenarios.

In contrast, our work integrates multiple techniques to provide a practical and robust solution for real-time control. By focusing on efficient and scalable methods, we ensure that our system can handle high-speed motions and maintain synchronization, even with simplified user setups (Fig. \ref{fig:intro}).

\subsection{Contributions}
Our contributions can be summarized as:
\begin{itemize}
    \item Minimal motion capture and retargeting framework, requiring only 7 IMUs to generate full-body references for the robot.
    \item High-speed whole-body streaming that uses a series of filtering, estimation, and prediction techniques, which allows us to achieve high control bandwidth (1kHz) even when the user references are acquired at a low rate (60Hz).  
     \item Integration of cycloidal actuators for high-bandwidth motion and impact resilience.
\end{itemize}

\begin{figure*}
    \centering
        \includegraphics[width=1.0\textwidth]{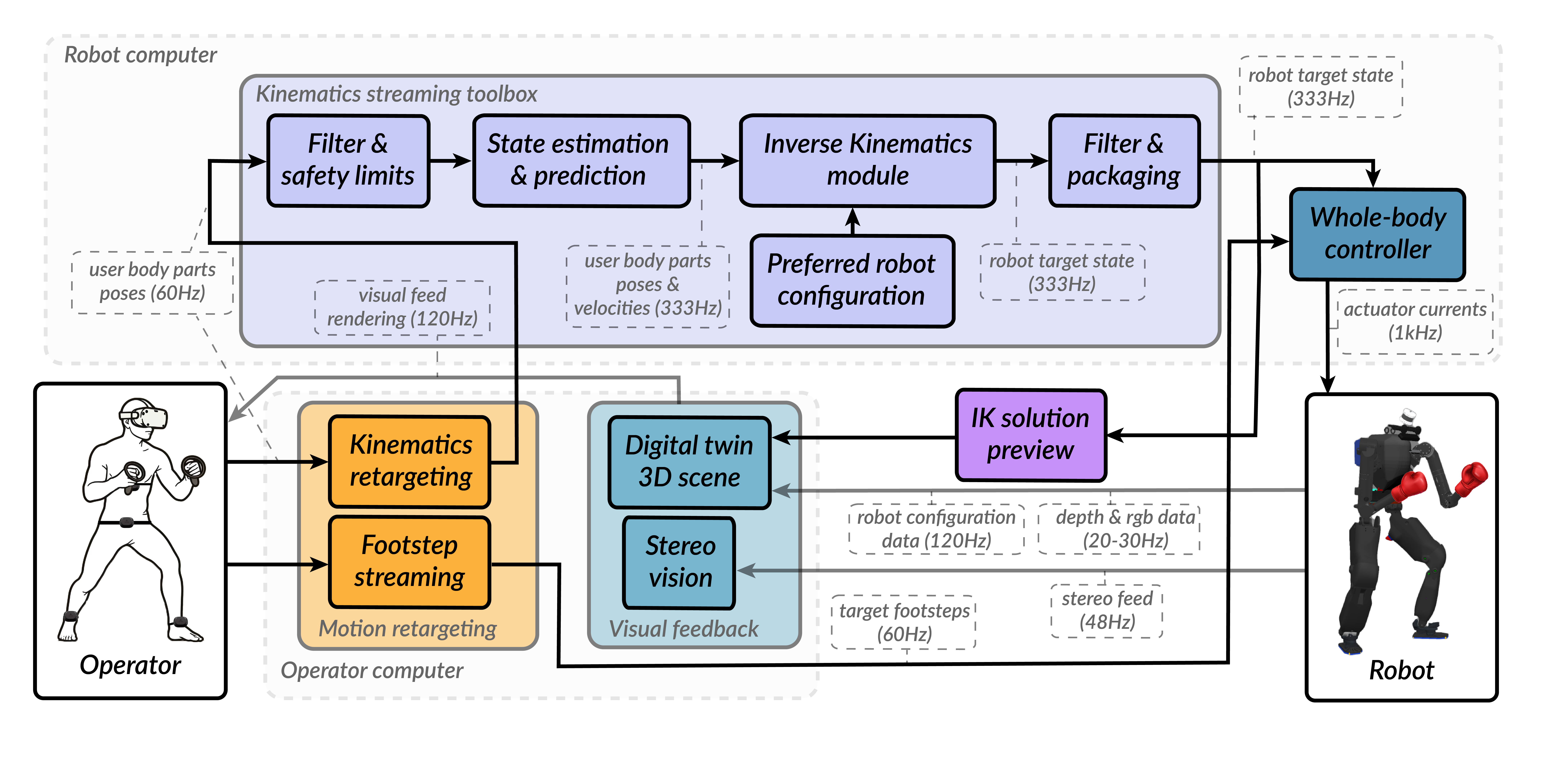}
    \caption{Flowchart of the proposed teleoperation system.}
    \label{fig:flowchart}
\end{figure*}

\section{Overview}
\label{overview}
The proposed teleoperation system integrates multiple components to achieve high-speed and seamless control. Fig. \ref{fig:flowchart} illustrates a flowchart of the system. 

The motion retargeting involves two key processes: footstep streaming and kinematics retargeting. Footstep streaming continuously tracks the user’s foot movements, estimating new footstep poses for the robot accordingly, whereas kinematics retargeting translates the operator’s upper body movements into corresponding robot movements. Using a VR headset and motion trackers, the system captures the operator's body poses and maps these to the robot’s references. This allows the robot to replicate the operator’s actions closely.

The Kinematics Streaming Toolbox (KST) then takes over, ensuring that the operator's commands are executed in real-time with high precision. The KST begins with filtering and adding safety limits, ensuring all incoming commands from the operator are safe and within operational parameters. Next, the state estimation and prediction module smooths and anticipates future states of the robot, compensating for delays to maintain continuous and fluid motion.

The inverse kinematics (IK) module within the KST calculates the necessary joint movements to achieve the desired rigid body positions and orientations. These movements are then refined through additional filtering and packaging processes, preparing them for execution by the robot’s whole-body controller.

Finally, the whole-body controller on the robot side tracks and executes the commands from the KST and the footstep streaming module, while maintaining the robot's balance.
% TODO Remove/simplify this section
For visual feedback, the system uses a digital twin 3D scene \cite{penco2024} and stereo vision \cite{behnke2023, schwarz2021}. The digital twin offers a real-time virtual representation of the robot and its environment, while the stereo vision system provides a more immervise first person view with enhanced depth perception. Note that this paper focuses on the main contributions of this work and as such, does not elaborate on the visual feedback as it mostly relies on the current state of the art.
The subsequent sections provide a more detailed description of each component.

\section{Motion Retargeting}
\label{retargeting}
To capture user motion, we employed a straightforward setup that includes the Vive Focus 3 VR headset, two VR controllers, and four Vive Ultimate trackers. These trackers are strategically placed around the sternum, waist, and ankles. The system is completely wireless, requires no start-up calibration, and does not need any physiological measurements from the user.

Building on this streamlined capture system, we developed a comprehensive whole-body retargeting method. This method translates human motion into equivalent robot movements for use in the KST. Our approach can be adapted to any motion capture system employing a total of seven IMUs.

The retargeting process involves several steps:
\subsubsection{Pelvis retargeting} The robot's initial pelvis position is recorded in the world frame. Similarly, the initial transform of the waist tracker in the world frame is captured. The vertical (z-axis) movement of the waist tracker is scaled based on the ratio of the robot’s pelvis height with extended legs to the initial user’s waist height.
\begin{equation}
    \Delta_{\text{pelvis}} = \frac{h_{\text{pelvis;R}}}{h_{\text{pelvis;H}}}.
\end{equation}
This ensures that the pelvis height in the robot reflects the user's movement proportionally.
As the user moves, the variation in the waist tracker's position and orientation from its initial state is computed:
\begin{equation}
    \mathbf{p}^{k}_{\text{pelvis;R}} = \mathbf{p}^{0}_{\text{pelvis;R}} + \Delta_{\text{pelvis}} (\mathbf{p}^{k}_{\text{pelvis;H}} - \mathbf{p}^{0}_{\text{pelvis;H}}),
\end{equation}
\begin{equation}
    \mathbf{R}^{k}_{\text{pelvis;R}} = \mathbf{R}^{0}_{\text{pelvis;R}} \mathbf{R}^{k}_{\text{pelvis;H}} \left( \mathbf{R}^{0}_{\text{pelvis;H}} \right)^{-1},
\end{equation}
where the superscripts $0$ and $k$ refer to measurements at initial time and at time $k$, and the subscripts $\text{H}$ and $\text{R}$ indicate measurements on human and robot, respectively.
To prevent unnatural motions, particularly during double support phases, roll variations are not taken into account.
The combined transform of the initial pelvis and the scaled waist variation is applied to update the robot's pelvis pose. 

\subsubsection{Hands retargeting} 
First we estimate the user's shoulder positions, utilizing the positions of the chest tracker and the headset. Typically in a human, the distance from the sternum to the top of the head is approximately equivalent to the length of two heads. Given that the headset is positioned around the midpoint of the head, the distance between the chest tracker and the headset is roughly 1.5 times the length of a head. This relationship allows us to calculate the head length \( l_{\text{head}} \) as:

\begin{equation}
    l_{\text{head}} = \frac{\|\mathbf{p}_{\text{head}} - \mathbf{p}_{\text{chest}}\|}{1.5}
\end{equation}

Using this head length, we can approximate the positions of the shoulders. The shoulders are estimated to be offset horizontally from the chest by a distance equal to the calculated head length, and vertically halfway between the chest tracker and the headset. Thus, the positions of the left (\( \mathbf{p}_{\text{shoulder,L}} \)) and right (\( \mathbf{p}_{\text{shoulder,R}} \)) shoulders are:

\begin{equation}
    \mathbf{p}_{\text{shoulder,L(R)}} = \mathbf{p}_{\text{chest}} + \begin{bmatrix} 0 \\ (-)l_{\text{head}} \\ \frac{(\mathbf{p}_{\text{head}}.z - \mathbf{p}_{\text{chest}}.z)}{2} \end{bmatrix}
\end{equation}
The shoulder orientation can be approximated as being aligned with the orientation provided by the chest tracker.

Next, we record the initial positions $l_{\text{arm;H}}$ of the user's hands (VR controllers) relative to their shoulders.
To ensure the robot's arm movements are proportional to the user's, we scale this length to match the robot's arm length ($l_{\text{arm,R}}$). The scaling factor is:

\begin{equation}
\Delta_{\text{arm}} = \frac{l_{\text{arm,R}}}{l_{\text{arm,H}}}
\end{equation}

As the user moves their hands, we apply this scaling factor to adjust the robot's hand positions. The updated positions of the robot's hands are computed as follows:

\begin{equation}
\mathbf{p}_{\text{hand;R}}^{\text{updated}} = \mathbf{p}_{\text{shoulder;R}} + \Delta_{\text{arm}} (\mathbf{p}_{\text{hand;H}} - \mathbf{p}_{\text{shoulder;H}})
\end{equation}

This ensures that the robot's hand movements are a scaled reflection of the user's hand movements. Furthermore, the orientation of the hands matches the VR controllers' orientation.

\subsubsection{Center of mass retargeting} 
To effectively transfer the human’s center of mass (CoM) to the robot, we use a method based on normalized offsets, allowing us to reconstruct the robot's CoM ground position from the human’s data \cite{penco2018robust}. 
To approximate the user's CoM, we start by projecting the waist tracker onto the ground plane ($\mathbf{p}_{\text{g,waist;H}}$).
Next, we determine the ground positions of the user's left and right ankles using the data from the ankle trackers. These positions provide a stable reference for the user's feet on the ground ($\mathbf{p}_{\text{g,lFoot;H}}$, $\mathbf{p}_{\text{g,rFoot;H}}$).
This position is calculated relative to the left foot, and then projected onto the line connecting the left and right feet.
To standardize this position, we compute a normalized offset \( o \) that lies within the range [0, 1]. The offset \( o \) is given by:

\begin{equation}
    o = \frac{(\mathbf{p}_{\text{g,waist;H}} - \mathbf{p}_{\text{g,lFoot;H}}) \cdot (\mathbf{p}_{\text{g,rFoot;H}} - \mathbf{p}_{\text{g,lFoot;H}})}{\|\mathbf{p}_{\text{g,rFoot;H}} - \mathbf{p}_{\text{g,lFoot;H}}\|^2}.
\end{equation}

Using this calculated offset, we reconstruct the robot's CoM ground position along the line connecting the robot's feet. This reconstructed position is given by:

\begin{equation}
\mathbf{p}_{\text{g,CoM;R}} = \mathbf{p}_{\text{g,lFoot;R}} + o (\mathbf{p}_{\text{g,rFoot;R}} - \mathbf{p}_{\text{g,lFoot;R}}),
\end{equation}
where \( \mathbf{p}_{\text{g,lFoot;R}} \) and \( \mathbf{p}_{\text{g,rFoot;R}} \) represent the ground positions of the robot's left and right feet, respectively. 

\subsubsection{Footstep streaming} 
The footstep streaming algorithm operates continuously, monitoring the poses of the ankle trackers. For each foot, the system assesses whether the user is stepping by comparing the current tracker position to its initial position. If the foot has moved significantly—exceeding a predefined step threshold—and has been lifted beyond a specific lift threshold, the system identifies this as a step.

Once these conditions are satisfied, the system proceeds to estimate the final footstep position. First, the direction of movement is normalized:
\begin{equation}
    \mathbf{p}_{\text{distance, norm}} = \frac{\mathbf{p}_{\text{ankle, current}} - \mathbf{p}_{\text{ankle, initial}}}{\|\mathbf{p}_{\text{ankle, current}} - \mathbf{p}_{\text{ankle, initial}}\|}
\end{equation}

Then, the stride vector is calculated by scaling this normalized direction by a predefined stride length $\Delta_{\text{stride}}$:

\begin{equation}
\mathbf{p}_{\text{stride}} = \mathbf{p}_{\text{distance, norm}} \Delta_{\text{stride}}
\end{equation}

The new footstep position (\( \mathbf{p}_{\text{footstep}} \)) is determined by applying this stride vector to the current position of the robot’s foot (\( \mathbf{p}_{\text{foot;R}} \)):

\begin{equation}
\mathbf{p}_{\text{footstep}} = \mathbf{p}_{\text{foot;R}} + \mathbf{p}_{\text{stride}}
\end{equation}

In addition to positioning, the algorithm adjusts the yaw, or rotation around the vertical axis, based on changes in the tracker's yaw orientation. If the change in yaw exceeds a predefined turning threshold, the yaw of the footstep is adjusted accordingly to ensure it aligns with the user’s intended direction. This yaw adjustment ensures the footstep is correctly oriented with the user's movement direction and can initiate an in-place step whenever a change in footstep orientation is necessary.

After determining the final position and orientation of the footstep, the system checks if the computed pose is kinematically feasible. If it is, this information is sent to the walking controller for execution (Section \ref{sec:controller}).

After placing a footstep, the system continuously monitors the tracker's position to ensure stability. If the tracker's movement remains within a predefined stability threshold  over a sufficient number of iterations, the system resets the stepping state for that side. This allows the user to initiate a new step. This process prevents the prediction of an additional step when the user's leg is swinging.

\section{Kinematics streaming toolbox}
\label{streaming}
The KST is designed to enable low-latency whole-body motion streaming. It allows operators to safely control rapid robot movements in real-time through arbitrary motion-capture interfaces. Its design is focused on ensuring precise and timely execution of commands, which is critical in fast-paced teleoperation environments.

One of the key advantages of using a toolbox like the KST is the flexibility it provides in the choice of the framework used to generate input for the kinematics streaming. By decoupling the input generation from the execution, the toolbox can integrate with various motion-capture and control systems, allowing it to adapt to different teleoperation setups and technologies.

Moreover, the toolbox is designed to ensure reliable execution frequencies by operating in a separate thread. This threading strategy guarantees that the timing of operations remains consistent, independent of other processes running on the system. As a result, the toolbox can maintain a stable update rate, which is essential for smooth and responsive control.

Running the toolbox on the robot side further enhances its reliability. By processing the input messages and generating the output trajectories locally on the robot, the system can minimize the impact of network-induced delays, ensuring the timing between computed trajectory set-points is more predictable and consistent.

%\subsection{Toolbox API}
%The toolbox API is built to handle detailed control of the robot’s movements through a series of messages using ROS2.
%The API accommodates several types of input messages:
%\begin{itemize}
%    \item \textbf{Session control messages} manage the toolbox's operational state by starting, ending, or reinitializing a session;
%    \item  \textbf{Motion input messages} specify which body parts to track and their desired poses, directing how the robot's whole-body will move;
%     \item Additional \textbf{toolbox configuration messages} allow for customization of the toolbox’s behavior, for example enabling/disabling collision avoidance, or locking certain rigid body poses.
%\end{itemize}
% 
%On the output side, the toolbox generates:
%\begin{itemize}
%\item A \textbf{robot configuration message} that conveys the real-time %configuration computed by the solver, enabling continuous monitoring;
%\item \textbf{Whole-Body trajectory execution messages} that are streamed directly %to the robot controller for real-time execution of whole-body trajectories.
%\end{itemize}

\subsection{Filter and safety limits}
Several filtering components in the KST ensure that all kinematic inputs, specified through motion input messages, are within safe and operationally feasible parameters.

\subsubsection{Bounding box}
The bounding box constraint restricts the robot’s commands to a specified 3D volume relative to the robot's mid-feet reference frame. This frame is updated dynamically to align with the robot’s current stance, ensuring that the bounding box moves with the robot.
\subsubsection{Rate of change and velocity limits}
The KST imposes rate of change and velocity limits to prevent abrupt or unsafe movements. Any input that exceeds these thresholds is invalidated to maintain stability and safety.

\subsection{State estimation and prediction}
The KST integrates state estimation and prediction mechanisms to provide smooth and responsive desired inputs to be used as objectives in the inverse kinematics QP solver. The KST employs state estimation to determine the current pose and velocity of the kinematics inputs, e.g. user body parts, with the goal of smoothing out delayed or jittery signals.
Two types of state estimation are supported:
\subsubsection{First-order estimation}
This method first estimate the input velocities using a first-order finite difference on the input position and orientation.
Subsequently, these velocities are used for a first-order extrapolation to predict future states.
This approach ensures that if no new input is received in the next control cycle, the system can continue to predict and maintain smooth motion.
Whenever a new input is available the linear velocity $\mathbf{v}$ and the angular velocity $\omega$ are updated as follows:
\begin{equation}
\mathbf{v}_{\text{fd}} = \frac{\mathbf{p}_{\text{current}} - \mathbf{p}_{\text{previous}}}{dt}
\end{equation}
\begin{equation}
\boldsymbol{\omega}_{\text{fd}} = \frac{1}{dt} log\left(\mathbf{q}^{-1}_\text{previous} \mathbf{q}_\text{current} \right)
\end{equation}
where $\mathbf{p}$ is input position, $\mathbf{q}$ the input quaternion, and $dt$ is the period of the toolbox.

When no new input is available, the velocity is decayed towards zero using linear interpolation over a fixed duration.
This ensures smooth and continuous motion even if the input is momentarily unavailable. 
The resulting estimated velocity that is used in the first-order extrapolation can be written as:
\begin{equation}
\mathbf{v}_{\text{estimated}} = \alpha(t) \mathbf{v}_{\text{fd}}
\end{equation}
\begin{equation}
\mathbf{\omega}_{\text{estimated}} = \alpha(t) \mathbf{\omega}_{\text{fd}}
\end{equation}
where \( \alpha \)(t) is the linear decay function starting from 1 when a new input is received, linearly decreasing to 0 over a fixed user-determined duration.
When \( \alpha \)(t) reaches 0, the KST assumes that the corresponding end-effector is no longer controlled.

\subsubsection{Feedback controller-based estimation}
Unlike the above method which may result in discontinuities in the estimated positions and orientations, the feedback controller approach focuses on maintaining continuity in the estimated pose.
Similar to the previous approach, it uses a first-order finite difference to estimate the input velocity onto which it adds a correction term computed from the error in pose.
The correction term is computed to linearly cancel out the error measured on each update over a small duration $T_\text{corr}$ during the prediction phase:
\begin{equation}
\mathbf{v}_{\text{corr}} = \frac{\mathbf{p}_{\text{current}} - \mathbf{p}_{\text{estimated}}}{T_{corr}}
\end{equation}
\begin{equation}
\boldsymbol{\mathbf{\omega}}_\text{corr} = \frac{1} {T_{corr}} log(\mathbf{q}^{-1}_\text{estimated} \mathbf{q}_\text{current})
\end{equation}
Similar to the previous approach, the future state is predicted using a first-order integration and uses the same decay process.
Each time a new input is available, the input velocity and the correction term are updated.
The resulting estimated velocity that is used in the first-order extrapolation can be written as:
\begin{equation}
\mathbf{v}_{\text{estimated}} = \alpha(t) (\mathbf{v}_{\text{fd}} + \mathbf{v}_{\text{corr}})
\end{equation}
\begin{equation}
\mathbf{\omega}_{\text{estimated}} = \alpha(t) (\mathbf{\omega}_{\text{fd}} + \mathbf{\omega}_{\text{corr}})
\end{equation}

\subsection{Inverse Kinematics module}
Given the filtered input, the inverse kinematics (IK) module aims at solving for the robot joint velocities that will provide high tracking responsiveness as well as safe and smooth execution. Each update tick, the IK solves the following quadratic program (QP):

\begin{equation} \label{eq:IKQP}
\begin{aligned}
    \min_{\mathbf{v}_d} \quad   & c_{\mathrm{nom}} + c_{\mathbf{J}} + c_{\mathbf{v}_d}    \\
    \textrm{s.t.} \quad         & \mathbf{v}_{min} \leq \mathbf{v}_d \leq \mathbf{v}_{max}      \\
    \quad                       & \mathbf{A}\mathbf{v}_d \leq \mathbf{H} \\
    \quad                       & \mathbf{J}_{coll}\mathbf{v}_d \leq \mathbf{p}_{coll}
\end{aligned}
\end{equation}

The objective function terms are given by: \vspace{1mm}

\renewcommand{\arraystretch}{1.35}
\setlength{\tabcolsep}{2.5pt}

\begin{tabular}{ll}
 Nominal Objective: & $c_{\mathrm{nom}} = (\mathbf{v}_d - \mathbf{v}_{\mathrm{nom}})^T  \mathbf{C}_{\mathrm{nom}}(\mathbf{v}_d - \mathbf{v}_{\mathrm{nom}}) $ \\
 Kinematic Objective: & $c_{\mathbf{J}} = (\mathbf{J}\mathbf{v}_d - \mathbf{p})^T       \mathbf{C}_{\mathbf{J}} (\mathbf{J}\mathbf{v}_d - \mathbf{p}) $ \\
 Velocity Cost: & $c_{\mathbf{v}_d} = \mathbf{v}_d^T \mathbf{C}_{\mathbf{v}_d} \mathbf{v}_d$
\end{tabular}
% Need to force extra space, otherwise it's pretty tight
\\

Where the terms are given by:
\begin{itemize}
    \item $\mathbf{v}_d$ is the vector of desired joint velocities % and $\mathbf{q}$ is the corresponding vector of joint positions.
    \item $\mathbf{v}_{\mathrm{nom}}$ drives the robot to a nominal whole-body configuration.
    \item $\mathbf{J} = [\mathbf{J}^T_1 \ldots \mathbf{J}^T_k]^T$ and $\mathbf{p} = [\mathbf{p}^T_1 \ldots \mathbf{p}^T_k]^T$ are the stacked Jacobian matrices and motion objectives, see below for additional details.
    \item $\mathbf{v}_{min}$ and $\mathbf{v}_{max}$ bound the joint velocity. A nominal set of bounds are used unless a joint is near a limit, in which case the velocity is constrained such that the joint remains within its range of motion within the update tick $\Delta T$.
    \item $\mathbf{A}$ is the linear centroidal momentum matrix  \cite{orin2008centroidal} and $\mathbf{h}=\mathbf{A}\mathbf{v}_d$ is the linear momentum. $\mathbf{H}$ constrains the centroidal momentum to such that the CoM remains inside the support region.
    \item $\mathbf{J}_{coll} = [\mathbf{J}^T_{coll,1} \ldots \mathbf{J}^T_{coll,k}]^T$ and $\mathbf{p}_{coll} = [\mathbf{p}^T_{coll,1} \ldots \mathbf{p}^T_{coll,k}]^T$ are the stacked Jacobian matrices and velocity limits used to resolve self-collisions. Simple geometric primitives (spheres and capsules) are used for efficient representation of the robot collision model. Each update tick, the list of potential collisions, according to a predetermined threshold on minimum separation distance, are evaluated and collected. For each potential collision, we evaluate the pair of closest points and the collision axis. We then express the geometric Jacobian at one of the collision points and project it along the collision axis. Finally, we evaluate the maximum relative velocity along the collision axis that ensures collision is avoided for the next tick.
    \item $\mathbf{C}_{\mathrm{nom}}$, $\mathbf{C}_{\mathbf{J}}$, and $\mathbf{C}_{\mathbf{v}_d}$ are positive diagonal weight matrices.
\end{itemize}
For the kinematic objectives, each motion task is given by:
\begin{equation}
    \mathbf{J}_i \mathbf{v}_d = \mathbf{p}_i, \forall i\in[1;k]
\end{equation}
A motion objective $\mathbf{p}_i$ is computed from a feedback controller to track a reference. The reference can be a desired input for tracking the user in VR, or "zero" to minimize the controlled quantity, e.g. angular momentum. Kinematics tasks can be:
\begin{itemize}
    %\item Joint position, in which $\mathbf{J}_i$ is a selection matrix for the joint and $\mathbf{p}_i$ is a velocity towards a desired position. This type of task was not used in these experiments but can be used to hold an uncontrolled limb in place allowing the user to disable/enable tracking of a hand at runtime for instance.
    \item CoM of the robot, in which a desired CoM position in world is specified. The Jacobian is the linear centroidal momentum matrix $\mathbf{A}$ and the objective $\mathbf{p}_i$ drives towards the desired CoM position. The desired CoM is either a predefined constant position with respect to the feet to strengthen balance, or the desired user CoM to track.
    \item Spatial pose of a rigid body, in which a desired position and orientation in world and corresponding frame fixed to the rigid body are specified. A proportional feedback law on the error transform is used to compute $\mathbf{p}_i$ \cite{bullo1995proportional}. This type of task is used to track the user's hand poses, chest orientation, and pelvis height and orientation.
    \item Momentum of the robot, in which the desired momentum is zero to minimize the resulting momentum. The Jacobian is the centroidal momentum matrix. This task is really useful to prioritize the use of the joints which have a lesser impact on momentum.
\end{itemize}
Each iteration, the solution $\mathbf{v}_d$ from the QP is integrated to update the desired robot state $\mathbf{q}$:
\begin{equation} \label{ikupdate}
    \mathbf{q} \leftarrow \mathbf{q} + \mathbf{v}_d \Delta T
\end{equation}

\subsection{Post-Processing}
Finally, the output of the IK module is post-processed to ensure continuity of the desireds as well as adding a few last control knobs to allow tweaking the responsiveness versus smoothness of the execution before sending the desireds to the real-time controller.
\subsubsection{Velocity Down-scaling} This first processing permits to dampen the execution for a smoother execution by down-scaling the desired velocity.
\subsubsection{Estimating Joint Accelerations} A new set of position $q_\text{fb}$, velocity $\dot{q}_\text{fb}$, and acceleration $\ddot{q}_\text{fb}$ are estimated for each 1-DoF joint.
We use a PD-control to compute the acceleration required to track the IK module outputs $q_\text{ik}$ and $\dot{q}_\text{ik}$:
\begin{equation}
\ddot{q}_\text{fb} = k_p (q_\text{ik} - q_\text{fb}) + k_d (\dot{q}_\text{ik} - \dot{q}_\text{fb})
\end{equation}
where $k_p$ and $k_d$ are user-defined gains. The acceleration $\ddot{q}_\text{fb}$ is then integrated once to update the velocity $\dot{q}_\text{fb}$ and a second time to update the position $q_\text{fb}$.
 A similar strategy is used for the floating base using feedback controller in SE(3).
This processor allows to estimate the desired acceleration required for each joint which is used in the controller to improve tracking as well as to improve continuity over time.
\subsubsection{Low-Pass Filtering} This processing provides another control knob to adjust the smoothness of the IK outputs at the cost of minor delays in execution.
\subsubsection{Initial Blending} This manages the start of the stream accounting for the initial discrepancy between the IK solution and the actual robot configuration. 

\section{Whole-body Controller}
\label{sec:controller}
The robot uses a momentum-based whole-body controller that is framed as a quadratic program (QP) \cite{koolen2016}. The controller's primary task is to track a desired rate of change of momentum, but it can simultaneously track a set of external motion objectives for the robot's pelvis height, chest orientation, end-effector poses and arm configurations. By solving the QP, the controller produces a joint acceleration vector and contact wrenches. These values are then used to calculate the desired actuator torques through inverse dynamics. 
For stepping we use the dynamic walking behavior described in \cite{koolen2016}.
It is comprised of a state machine with standing, transfer, and swing states.
Each walking state has associated motion tasks and active contacts that are used to achieve balance.
The controller is designed to execute a sequence of footstep poses specified by the user. To model balance dynamics, we employ the instantaneous capture point (ICP) concept \cite{pratt2006capturepoint}, which helps in generating foot pose, pelvis height, and pelvis orientation trajectories that align with the user’s desired foot placements while maintaining stability.

During the transfer phase, an ICP trajectory is generated, and the stance foot contact constraints are relaxed to allow the heel to lift, facilitating a more natural walking motion. The upper body remains stationary unless the user explicitly commands upper body movements. This ensures that the focus remains on achieving stable and controlled locomotion.

For a comprehensive understanding of the balance dynamics and control strategies employed, additional details can be found in \cite{Pratt2019_bookchapter}.

\section{Cycloidal Actuators for High-Speed and Impact-Resilient Operations}
\label{actuators}
\begin{figure}[!t]
    \centering
        \includegraphics[width=0.5\columnwidth]{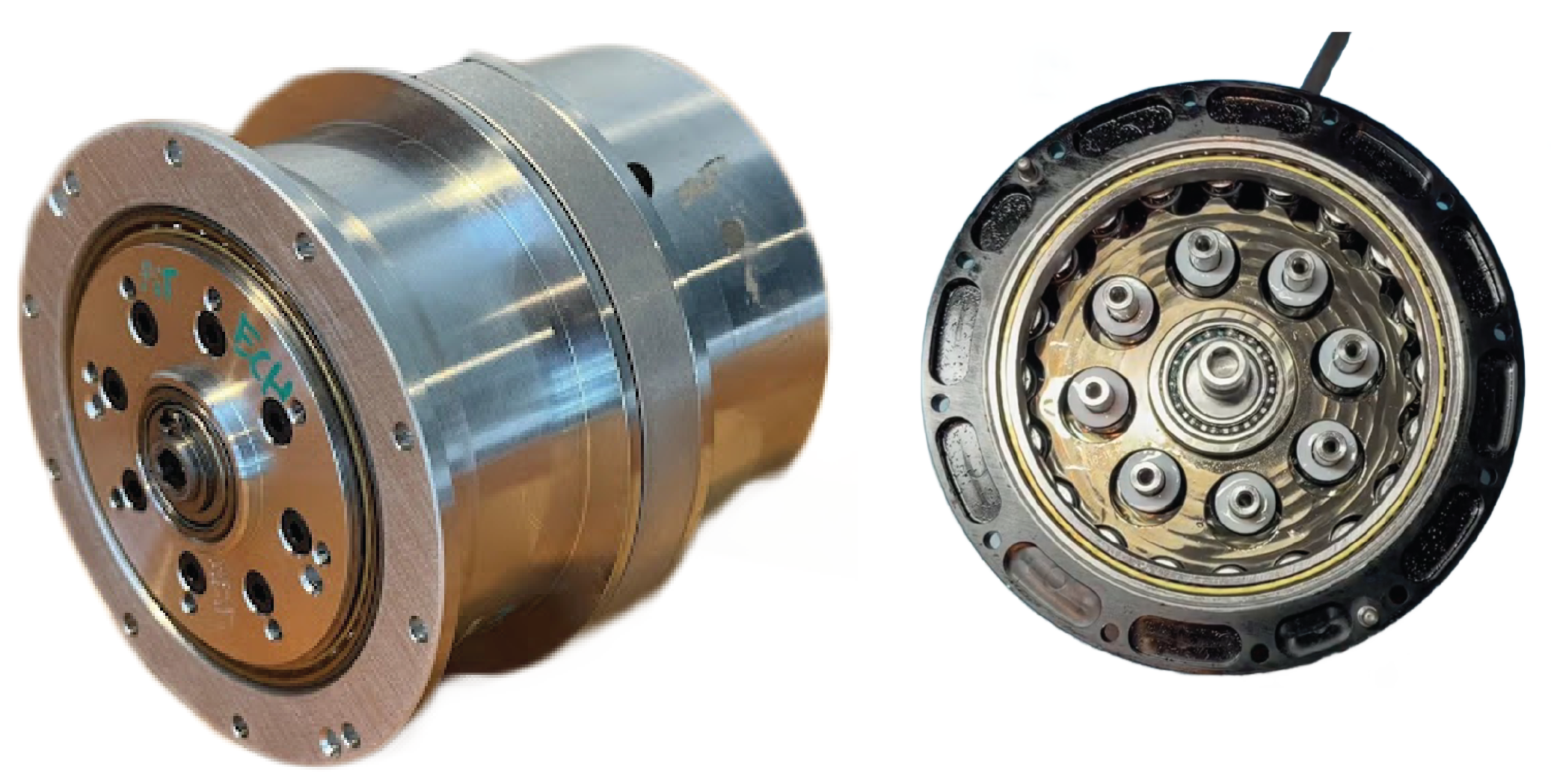}
    \caption{Nadia's cycloidal actuators and internal gear mechanism.}
    \label{fig:cycloidalActuator}
\end{figure}

\begin{figure}[!t]
    \centering
        \includegraphics[width=1.0\columnwidth]{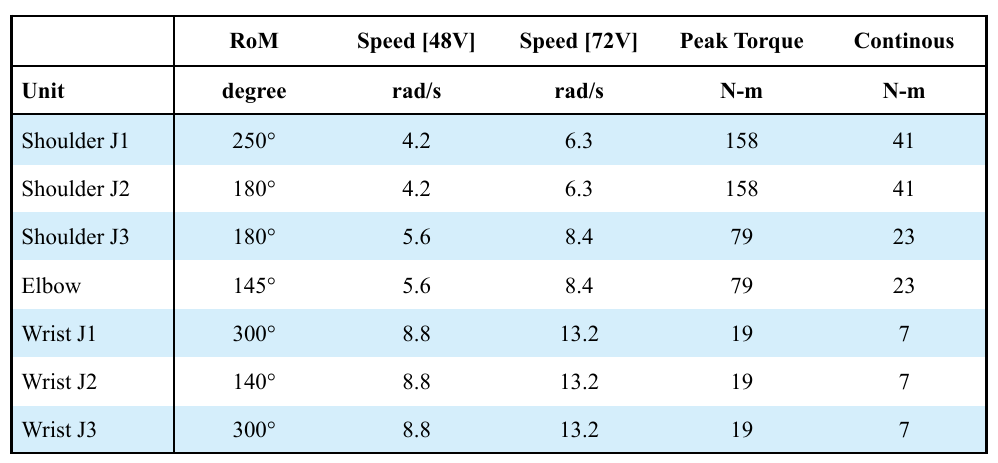}
    \caption{Performance specifications of cycloidal actuators in the robot's arms. The table details the range of motion (RoM), speed at 48V and 72V, peak torque, and continuous torque for each joint in the arm \cite{cycloidActuator2024a, cycloidActuator2024b}.}
    \label{fig:tab}
\end{figure}

A key component of our humanoid robot teleoperation system is the implementation of advanced cycloidal actuators  \cite{cycloidActuator2024a, cycloidActuator2024b} (Fig. \ref{fig:cycloidalActuator}). This type of actuation is known for its high efficiency and robustness \cite{Qi2024,slapak2022}, and provide the necessary torque and speed required for dynamic and responsive robot movements. Specifically, the cycloidal actuators in our design support a payload of up to 10kg and deliver a torque output up to 158 Nm (Fig. \ref{fig:tab}), ensuring the robot can perform a wide range of tasks with precision and strength.

Cycloidal gears have a unique design where the load is distributed across multiple contact points within the gear mechanism. This distribution reduces the stress on individual components, significantly enhancing the actuator's ability to withstand high-impact forces. Additionally, the smooth, rolling motion of the cycloidal gears minimizes backlash and increases durability \cite{farrell2018}, further contributing to their resistance to shock and impact. Other advantages include high torque capacity, and compact size compared to involute gearboxes.

The integration of these actuators represents a significant element for achieving high-speed, impact-resilient operations in humanoid robotics.

\section{Experiments and Results}
\label{results}
\begin{figure}
    \centering
    \includegraphics[width=1.0\linewidth]{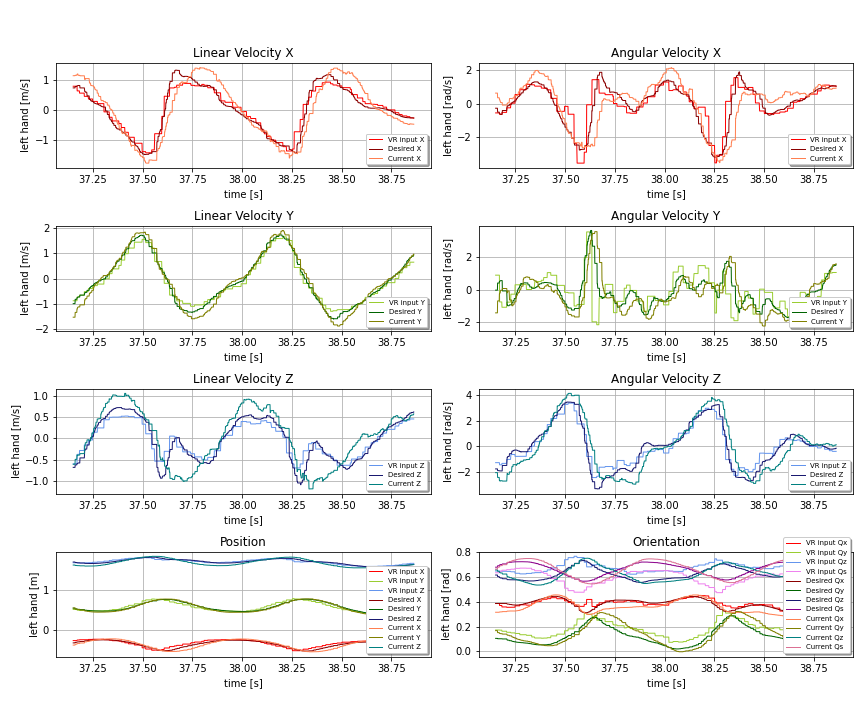}
    \caption{Comparison of VR input, desired, and current poses and velocity for the left robot hand during repeated left punches in free motion.}
    \label{fig:airPunch}
\end{figure}

\begin{figure}
    \centering
    \includegraphics[width=1.0\linewidth]{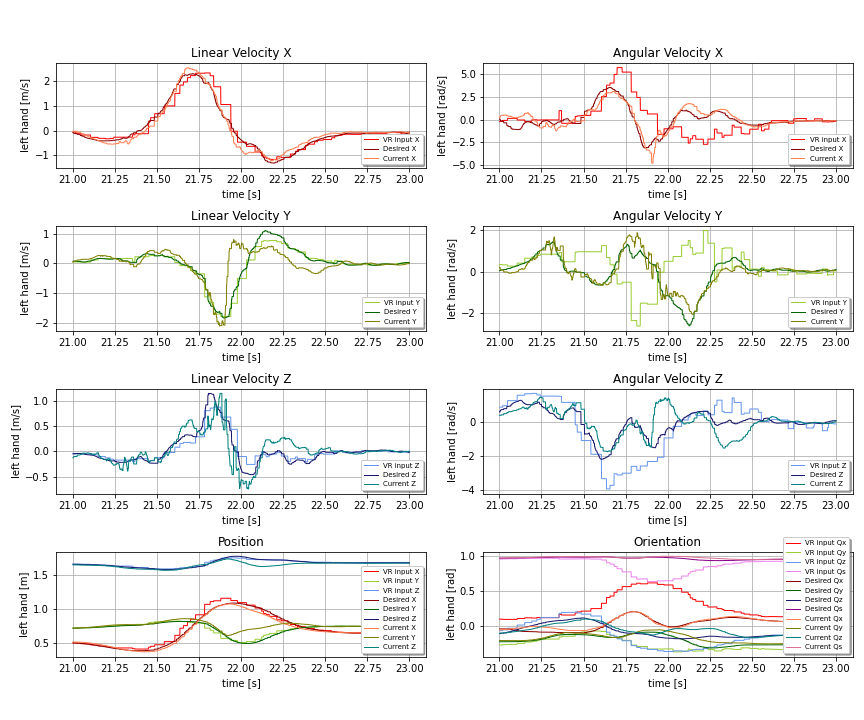}
    \caption{Comparison of VR input, desired, and current poses and velocity for the left robot hand during a left hook punch impacting a punching bag.}
    \label{fig:somePunches}
\end{figure}

\begin{figure*}
    \centering
    \includegraphics[width=1.0\linewidth]{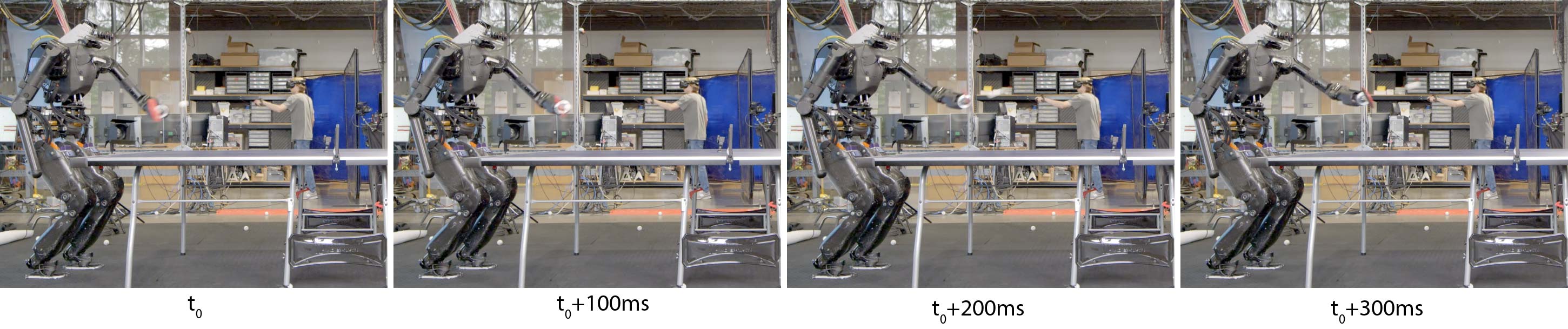}
    \caption{Screenshots of the teleoperation experiments showing the humanoid robot Nadia playing ping pong with a human opponent.}
    \label{fig:pingpong}
\end{figure*}

\begin{figure*}
    \centering
    \includegraphics[width=1.0\linewidth]{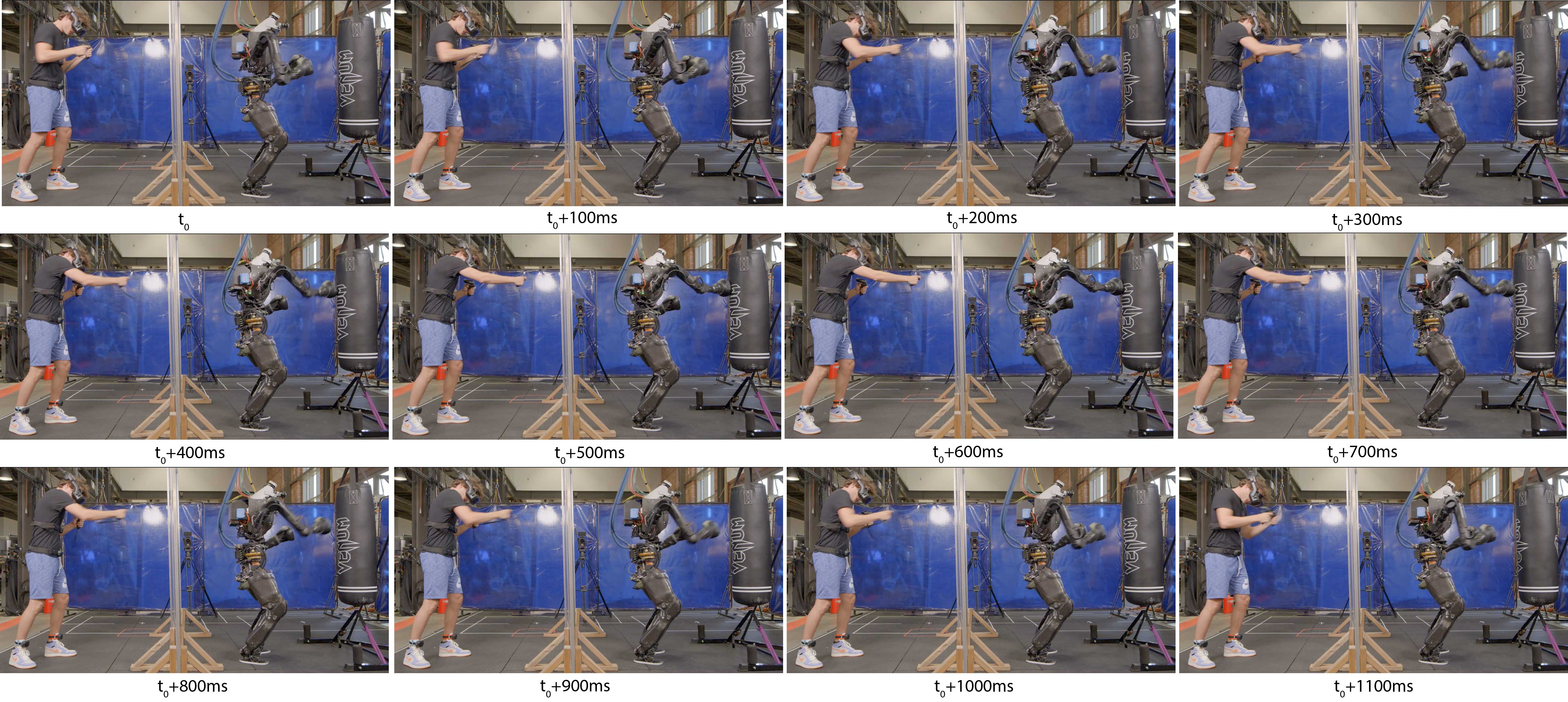}
    \caption{Screenshots of the teleoperation experiments showing the humanoid robot Nadia punching a 100lbs punching bag.}
    \label{fig:boxing}
\end{figure*}

We conducted multiple teleoperation experiments with our humanoid robot, Nadia. In one experiment, we teleoperated Nadia to play ping pong with a human opponent to demonstrate the synchronization capabilities of our approach (see Fig. \ref{fig:pingpong} and accompanying video). Ping pong was chosen to stress test the roundtrip latency of the proposed framework, from the user perceiving the ball's position to adjusting the racket's position and orientation.
To test the speed and transparency of our teleoperation system, we conducted experiments involving punching motions (Fig. \ref{fig:boxing}). Punching is a fast, whole-body motion that requires precise coordination. Additionally, the impact with the punching bag places significant stress on the motors and challenges the tracking of teleoperated references.
Fig. \ref{fig:airPunch} shows tracking data for the robot left hand while the operator is throwing is few straight punches without contact. It highlights that while the robot operates through a large range of velocities, we still have really good tracking performance. Note that the arms in this experiment only has four degrees of actuation, making it difficult to track all six degrees of freedom of the hand. The delay from the time the Kinematics Streaming Toolbox receives a new VR input, to the robot reaching the pose corresponding to that input is about 70 milliseconds.
Fig. \ref{fig:somePunches} shows the same tracking data but this time the operator is throwing a left hook and the robot makes contact with the punching bag. The instant of impact is especially noticeable in the y-axis position and linear velocity at the 21.9-second mark. The bouncing impact causes the hand to reverse motion rapidly, but tracking recovers in the following second.

\section{Conclusions}
\label{conclusions}
We presented an integrated solution for teleoperating a humanoid robot using a minimal and calibration-free user setup. Our framework utilizes filtering, estimation, and prediction techniques to enable high-speed motion, ensuring excellent transparency and synchronization between the user and robot movements. The integration of cycloidal actuators further enhances the robot's impact resilience. Thanks to these combined solutions, we have achieved, to the best of our knowledge, the fastest teleoperation experiments ever performed on a real humanoid robot.

\subsection{Source Code and Media}

The modules discussed in this paper can be accessed on our GitHub repository at \url{https://github.com/ihmcrobotics}.
The accompanying video can be found at \url{https://youtu.be/F6dqCauGPEM}.

% eventually replace this with generated final bbl maybe?
\bibliography{mybib}

\begin{thebibliography}{10}
\providecommand{\url}[1]{#1}
\csname url@rmstyle\endcsname
\providecommand{\newblock}{\relax}
\providecommand{\bibinfo}[2]{#2}
\providecommand\BIBentrySTDinterwordspacing{\spaceskip=0pt\relax}
\providecommand\BIBentryALTinterwordstretchfactor{4}
\providecommand\BIBentryALTinterwordspacing{\spaceskip=\fontdimen2\font plus
\BIBentryALTinterwordstretchfactor\fontdimen3\font minus \fontdimen4\font\relax}
\providecommand\BIBforeignlanguage[2]{{%
\expandafter\ifx\csname l@#1\endcsname\relax
\typeout{** WARNING: IEEEtran.bst: No hyphenation pattern has been}%
\typeout{** loaded for the language `#1'. Using the pattern for}%
\typeout{** the default language instead.}%
\else
\language=\csname l@#1\endcsname
\fi
#2}}

\bibitem{dallard2023}
A.~Dallard, M.~Benallegue, F.~Kanehiro, and A.~Kheddar, ``Synchronized human-humanoid motion imitation,'' \emph{IEEE Robotics and Automation Letters}, vol.~8, no.~7, pp. 4155--4162, 2023.

\bibitem{penco2023}
L.~Penco, J.-B. Mouret, and S.~Ivaldi, ``Prescient teleoperation of humanoid robots,'' in \emph{2023 IEEE-RAS 22nd International Conference on Humanoid Robots (Humanoids)}, 2023, pp. 1--8.

\bibitem{teleopSurvey}
K.~Darvish, L.~Penco, J.~Ramos, R.~Cisneros, J.~Pratt, E.~Yoshida, S.~Ivaldi, and D.~Pucci, ``Teleoperation of humanoid robots: A survey,'' \emph{IEEE Transactions on Robotics}, vol.~39, no.~3, pp. 1706--1727, 2023.

\bibitem{ramos2018}
J.~Ramos and S.~Kim, ``Humanoid dynamic synchronization through whole-body bilateral feedback teleoperation,'' \emph{IEEE Transactions on Robotics}, vol.~34, no.~4, pp. 953--965, 2018.

\bibitem{colin2023}
G.~Colin, J.~Byrnes, Y.~Sim, P.~M. Wensing, and J.~Ramos, ``Whole-body dynamic telelocomotion: A step-to-step dynamics approach to human walking reference generation,'' in \emph{2023 IEEE-RAS 22nd International Conference on Humanoid Robots (Humanoids)}, 2023, pp. 1--8.

\bibitem{ishiguro2020}
Y.~Ishiguro, T.~Makabe, Y.~Nagamatsu, Y.~Kojio, K.~Kojima, F.~Sugai, Y.~Kakiuchi, K.~Okada, and M.~Inaba, ``Bilateral humanoid teleoperation system using whole-body exoskeleton cockpit tablis,'' \emph{IEEE Robotics and Automation Letters}, vol.~5, no.~4, pp. 6419--6426, 2020.

\bibitem{fu2024humanplus}
Z.~Fu, Q.~Zhao, Q.~Wu, G.~Wetzstein, and C.~Finn, ``Humanplus: Humanoid shadowing and imitation from humans,'' in \emph{arXiv}, 2024.

\bibitem{he2024learning}
T.~He, Z.~Luo, W.~Xiao, C.~Zhang, K.~Kitani, C.~Liu, and G.~Shi, ``Learning human-to-humanoid real-time whole-body teleoperation,'' 2024.

\bibitem{penco2024}
L.~Penco, K.~Momose, S.~McCrory, D.~Anderson, N.~Kitchel, D.~Calvert, and R.~J. Griffin, ``Mixed reality teleoperation assistance for direct control of humanoids,'' \emph{IEEE Robotics and Automation Letters}, vol.~9, no.~2, pp. 1937--1944, 2024.

\bibitem{behnke2023}
M.~Schwarz, C.~Lenz, R.~Memmesheimer, B.~Pätzold, A.~Rochow, M.~Schreiber, and S.~Behnke, ``Robust immersive telepresence and mobile telemanipulation: Nimbro wins ana avatar xprize finals,'' in \emph{2023 IEEE-RAS 22nd International Conference on Humanoid Robots (Humanoids)}, 2023, pp. 1--8.

\bibitem{schwarz2021}
M.~Schwarz and S.~Behnke, ``Low-latency immersive 6d televisualization with spherical rendering,'' in \emph{2020 IEEE-RAS 20th International Conference on Humanoid Robots (Humanoids)}, 2021, pp. 320--325.

\bibitem{penco2018robust}
L.~Penco, B.~Cl{\'e}ment, V.~Modugno, E.~M. Hoffman, G.~Nava, D.~Pucci, N.~G. Tsagarakis, J.-B. Mouret, and S.~Ivaldi, ``Robust real-time whole-body motion retargeting from human to humanoid,'' in \emph{IEEE/RAS Int. Conf. on Humanoid Robots}, Nov 2018.

\bibitem{orin2008centroidal}
D.~E. Orin and A.~Goswami, ``Centroidal momentum matrix of a humanoid robot: Structure and properties,'' in \emph{2008 IEEE/RSJ International Conference on Intelligent Robots and Systems}.\hskip 1em plus 0.5em minus 0.4em\relax IEEE, 2008, pp. 653--659.

\bibitem{bullo1995proportional}
F.~Bullo and R.~M. Murray, ``Proportional derivative (pd) control on the euclidean group,'' 1995.

\bibitem{koolen2016}
T.~Koolen, S.~Bertrand, G.~Thomas, T.~de~Boer, T.~Wu, J.~Smith, J.~Englsberger, and J.~Pratt, ``Design of a momentum-based control framework and application to the humanoid robot atlas,'' \emph{International Journal of Humanoid Robotics}, vol.~13, no.~01, p. 1650007, 2016.

\bibitem{pratt2006capturepoint}
J.~Pratt, J.~Carff, S.~Drakunov, and A.~Goswami, ``Capture point: A step toward humanoid push recovery,'' in \emph{2006 6th IEEE-RAS International Conference on Humanoid Robots}, 2006, pp. 200--207.

\bibitem{Pratt2019_bookchapter}
\BIBentryALTinterwordspacing
J.~E. Pratt, S.~Bertrand, and T.~Koolen, \emph{Stepping for Balance Maintenance Including Push-Recovery}.\hskip 1em plus 0.5em minus 0.4em\relax Dordrecht: Springer Netherlands, 2019, pp. 1419--1466. [Online]. Available: \url{https://doi.org/10.1007/978-94-007-6046-2_41}
\BIBentrySTDinterwordspacing

\bibitem{cycloidActuator2024a}
``System and method for compensating characteristics of actuator to improve torque control and performance,'' Patent US Provisional Application No. 63/669,606, 2024.

\bibitem{cycloidActuator2024b}
``Low friction rolling contact transmission for an actuator,'' Patent US Provisional Application No. 63/670,000, 2024.

\bibitem{Qi2024}
\BIBentryALTinterwordspacing
L.~Qi, D.~Yang, B.~Cao, Z.~Li, and H.~Liu, ``Design principle and numerical analysis for cycloidal drive considering clearance, deformation, and friction,'' \emph{Alexandria Engineering Journal}, vol.~91, pp. 403--418, 2024. [Online]. Available: \url{https://www.sciencedirect.com/science/article/pii/S1110016824001182}
\BIBentrySTDinterwordspacing

\bibitem{slapak2022}
\BIBentryALTinterwordspacing
V.~Šlapák, J.~Ivan, K.~Kyslan, M.~Hric, F.~Ďurovský, D.~Paulišin, and M.~Kočiško, ``Measurement and modelling of a cycloidal gearbox in actuator with permanent magnet synchronous machine,'' \emph{Machines}, vol.~10, no.~5, 2022. [Online]. Available: \url{https://www.mdpi.com/2075-1702/10/5/344}
\BIBentrySTDinterwordspacing

\bibitem{farrell2018}
\BIBentryALTinterwordspacing
L.~C. Farrell, J.~Holley, W.~Bluethmann, and M.~K. O’Malley, ``{Cycloidal Geartrain In-Use Efficiency Study},'' vol. Volume 5B: 42nd Mechanisms and Robotics Conference, p. V05BT07A034, 2018. [Online]. Available: \url{https://doi.org/10.1115/DETC2018-85275}
\BIBentrySTDinterwordspacing

\end{thebibliography}

\end{document}